\documentclass{article}
\usepackage{ijcai22}

\usepackage{times}
\usepackage{url}
\usepackage[hidelinks]{hyperref}
\usepackage[utf8]{inputenc}
\usepackage[small]{caption}
\usepackage{graphicx}
\usepackage{amsmath}
\usepackage{booktabs}
\usepackage{subcaption}
\usepackage{tabularx}
\usepackage{calc}
\usepackage{hyperref}
\newcommand\mybox[1]{\rotatebox{90}{%
    \parbox{\widthof{Combined\ }}{\raggedright #1}}}
    
\title{Feasibility of Inconspicuous GAN-generated Adversarial Patches against Object Detection}
\author{
Svetlana Pavlitskaya$^1$\and 
Bianca-Marina Codău$^2$ and 
J. Marius Zöllner$^{1,2}$
\affiliations
$^1$FZI Research Center for Information Technology\\
$^2$Karlsruhe Institute of Technology (KIT) 
\emails
pavlitskaya@fzi.de,
bianca.codau@student.kit.edu,
zoellner@fzi.de
}

\begin{document}

\maketitle

\begin{abstract}
  Standard approaches for adversarial patch generation lead to noisy conspicuous patterns, which are easily recognizable by humans. Recent research has proposed several approaches to generate naturalistic patches using generative adversarial networks (GANs), yet only a few of them were evaluated on the object detection use case. Moreover, the state of the art mostly focuses on suppressing a single large bounding box in input by overlapping it with the patch directly. Suppressing objects near the patch is a different, more complex task. In this work, we have evaluated the existing approaches to generate inconspicuous patches. We have adapted methods, originally developed for different computer vision tasks, to the object detection use case with YOLOv3 and the COCO dataset. We have evaluated two approaches to generate naturalistic patches: by incorporating patch generation into the GAN training process and by using the pretrained GAN. For both cases, we have assessed a trade-off between performance and naturalistic patch appearance. Our experiments have shown, that using a pre-trained GAN helps to gain realistic-looking patches while preserving the performance similar to conventional adversarial patches.
\end{abstract}

\section{Introduction}

Deep neural networks (DNNs) are vulnerable to adversarial attacks in which input data is deliberately modified~\cite{szegedy2013intriguing}. In case of image data, adversarial noise is added to an input sample, affecting the entire image. Another type of attack is an adversarial patch, which can be positioned arbitrarily in a restricted region of an image. Patches can be applied to the input images digitally as well as in a real-world setting. But state-of-the-art research focuses on creating adversarial patches which are easily recognizable by the human eye. These are characterized by chaotic patterns, bright colors and do not resemble real-life objects but rather random noise ~\cite{brown2017adversarial,thys2019fooling,pavlitskaya2020feasibility}. A much harder problem is posed by creating inconspicuous patches as their purpose is to elude human detection while still being a threat to DNNs.

Recently, methods to enforce realistic appearance of adversarial patches have been proposed \cite{hu2021naturalistic,kong2020physgan,doan2021tnt}. Existing approaches aim at deterring image classifiers or steering models as well as object detectors. In the latter case, however, an adversarial patch manages to attack only one large object in an input image.

\begin{figure}
	\centering
	\begin{subfigure}[t]{0.48\linewidth}
    \includegraphics[width=\textwidth]{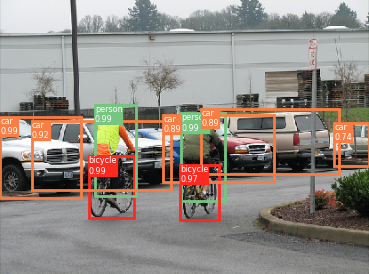}
    \caption{No attack}
    \end{subfigure}
    \begin{subfigure}[t]{0.48\linewidth}
    \includegraphics[width=\textwidth]{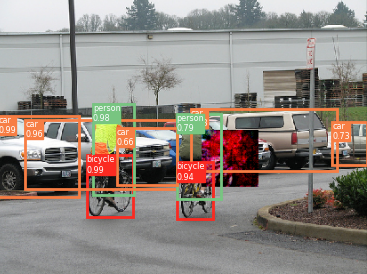}
    \caption{Pretrained DCGAN with patch transformations applied}
    \end{subfigure}
    \begin{subfigure}[t]{0.48\linewidth}
    \includegraphics[width=\textwidth]{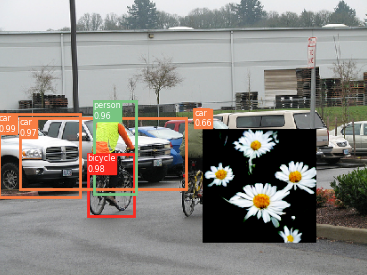}
    \caption{Pretrained BigGAN with latent shift applied}
    \end{subfigure}
    \begin{subfigure}[t]{0.48\linewidth}
    \includegraphics[width=\textwidth]{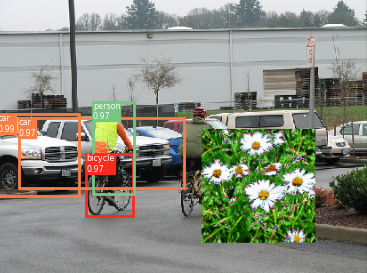}
    \caption{Pretrained BigGAN with patch transformations applied}
    \end{subfigure}
    \caption{Overview of the patches generated with the evaluated methods}
	\label{concept_pic}
\end{figure}

In this work, we perform extensive literature research and identify promising approaches to generate inconspicuous adversarial patches. We further apply these methods to the object detection use case. Differently from the existing work on naturalistic patches against object detection, the focus of our work is to affect objects in the attacked image, which are located near the patch. We run experiments in a digital setting in per-instance and universal manner. We further analyse which approach is the most suitable for the selected setting and discuss the trade-off between attack success and realistic appearance.

\section{Related Work}

\subsection{Conspicuous Adversarial Patches}

The idea of an adversarial perturbation restricted to a specific image area was first proposed by Brown et al. \cite{brown2017adversarial}. The first approaches focused on the image classification use case \cite{karmon2018laVAN}. Later, patch-based attacks for object detection were also proposed \cite{liu2018dpatch,thys2019fooling,lee2019physical}. A general approach consists in either maximizing the detector loss or, in case of an object vanishing attack, minimizing the detector loss for the empty label~\cite{chow2020targeted}. 

\begin{figure}[t]
	\centering
    \begin{subfigure}[t]{.18\linewidth}
    	\includegraphics[width=\textwidth]{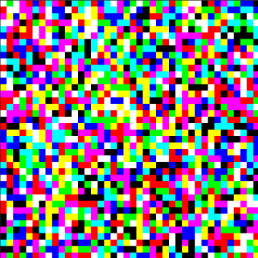}
    	\caption{\mybox{\cite{liu2018dpatch}}}
    \end{subfigure}
    \begin{subfigure}[t]{.18\linewidth}
    	\includegraphics[width=\textwidth]{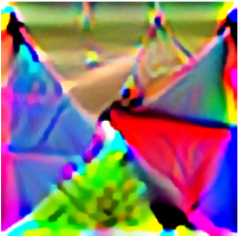}
    	\caption{\mybox{\cite{thys2019fooling}}}
    \end{subfigure}
    \begin{subfigure}[t]{.18\linewidth}
    	\includegraphics[width=\textwidth]{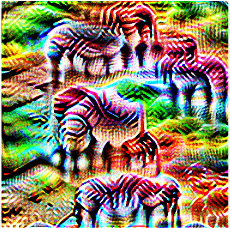}
    	\caption{\mybox{\cite{lee2019physical}}}
    \end{subfigure}
    \begin{subfigure}[t]{.18\linewidth}
    	\includegraphics[width=\textwidth]{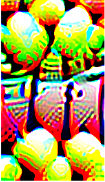}
    	\caption{\mybox{\cite{xu2020adversarial}}}
    \end{subfigure}
    \begin{subfigure}[t]{.18\linewidth}
    	\includegraphics[width=\textwidth]{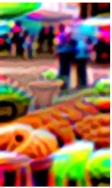}
    	\caption{\mybox{\cite{wu2020making}}}
    \end{subfigure}
    \caption{Examples of state-of-the-art \textbf{conspicuous} adversarial patches against object detection: (a-b) applied in a digital setting, (d-e) printed on a t-shirt}
    \label{fig:consp-sota}
\end{figure}

To enable attacks in the real-world setting, the non-printability loss component is usually added, which restricts pixel values to the set of printable colours. Furthermore, the total variation loss is usually applied in order to make colourful patterns of the generated adversarial patches appear smoother~\cite{mahmood2016accesorize}. The patches then be printed, e.g. on a t-shirt to fool object detectors Examples of adversarial patch attacks against object detection in the real world are \cite{xu2020adversarial,wu2020making}. Recently, a dataset of printable adversarial patches against object detection was introduced in \cite{braunegg2020apricot}. However, adversarial patches generated in the conventional way still have a conspicuous character (see Figure  \ref{fig:consp-sota}).

%An approach for generating adversarial patches which perform successful attacks in both digital and physical environments is presented by Thys et al. in \cite{thys2019fooling}. Thys et al. target their patch attacks on the class \emph{persons}, which is characterized by high variance in its individual instances. The objective is to minimize the loss, containing the non-printability score, the total variation loss and the maximum objectness score in the attacked image, which corresponds to the maximum probability that the bounding box contains an object. The non-printability score sums up the minimal differences between the pixel values in the patch and the colours in a given set of printable colours. The second loss is the total variation loss defined as the sum of differences between adjacent pixel values in the generated patch. By minimizing the total variation, the colorful patterns of the generated adversarial patches appear smoother, which makes them easier to use in a physical setting~\cite{mahmood2016accesorize}. However, this perturbation smoothing does not reduce the conspicuous character of the patches.

%The experiments performed by Thys et al. used the COCO dataset~\cite{thys2019fooling}. The model under attack was the YOLO version 2 object detector. The experiments show successful attacks when the patch is applied on the picture digitally as well as when the printed patch is filmed with a camera.

\subsection{Inconspicuous Adversarial Patches}

An inconspicuous adversarial patch can be enforced either by using a specific loss function or a generative adversarial network (GAN). The first group of approaches maximizes the loss function to obtain patches that resemble a certain real image. \textit{adv-watermark}~\cite{jia2020advwatermark}, for instance, generates adversarial patches as image watermarks by performing a heuristic random search for the global minimum as an adaptation of the \textit{Basin Hopping (BH)} optimization algorithm. 

Recently, approaches of the second group, which rely on  GANs, have gained popularity. We group GAN-based approaches into two categories: (1) methods which include patch generation directly into the GAN training process and (2) methods which generate an adversarial patch using a pretrained GAN.

A first attempt to use GANs to generate natural adversarial examples was performed by Zhao et al. \cite{zhao2018generating}. Here, a pretrained Wasserstein GAN \cite{arjovsky2017wasserstein} is combined with an inverter, which maps data to the latent representation. The experiments, however, were restricted to the image classification on MNIST and LSUN datasets as well as on a text generation task.

\begin{figure}[t]
	\centering
    \begin{subfigure}[t]{.32\linewidth}
    	\includegraphics[width=\textwidth]{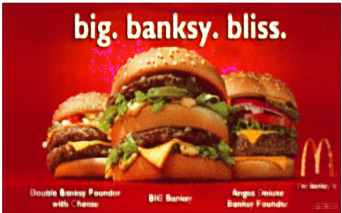}
    	\caption{PhysGAN \\\cite{kong2020physgan}}
    \end{subfigure}
    \begin{subfigure}[t]{.32\linewidth}
    	\includegraphics[width=\textwidth]{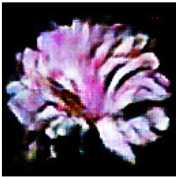}
    	\caption{TnT attack \\\cite{doan2021tnt}}
    \end{subfigure}
    \begin{subfigure}[t]{.32\linewidth}
    	\includegraphics[width=\textwidth]{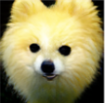}
    	\caption{Naturalistic \\\cite{hu2021naturalistic}}
    \end{subfigure}
    \caption{Examples of state-of-the-art \textbf{inconspicuous} adversarial patches against object detection}
    \label{fig:inconsp-sota}
\end{figure}

\subsubsection{Combined Patch-GAN Training}
 %The \textit{perceptual-sensitive GAN (PSGAN)} approach~\cite{DBLP:conf/aaai/LiuLFMZXT19} uses a GAN with the U-Net architecture to create patches that are then placed in image regions that have the highest impact on final predictions. 
 
 %losely related to PSGAN is the \textit{Inconspicuous Adversarial Patches (IAP)} framework~\cite{DBLP:journals/corr/abs-Bai}, which uses a series of GANs with different scales to create patches in a coarse-to-fine way improving their unobtrusive character. Most experiments on attacking the autonomous driving models focus on digitally applying patches whereas real-world attacks have been performed at a smaller scale~\cite{DBLP:conf/percom/DengZZCLK20}. 

\emph{PhysGAN} attack~\cite{kong2020physgan} is one representative of the first group of approaches. It is designed to generate patch attacks and place them in road side video footage to deter steering prediction models. For a given input video sequence, the algorithm learns a patch to be included into every frame. The PhysGAN model includes, next to a generator-discriminator pair, an encoder for extracting the features out of input video frames. The encoder output is then fed directly to the generator. The adversarial road sign, computed by the generator, and a real road sign are then sent to a discriminator. The resulting adversarial patch is then added to each frame of the original video sample creating an adversarial input. Finally, to obtain the perturbation, the generator is updated over the loss of the targeted model, calculated on the adversarial video slice, while taking the original frames as the ground truth. The resulting adversarial patch is indistinguishable from the roadside poster and leads to a noticeable prediction error.

Another approach designed to generate more realistic adversarial patches is the \emph{Perceptual-Sensitive GAN (PSGAN)}~\cite{liu2019psgan}. It was evaluated on the traffic sign recognition as well as on general image classification use cases. It adapts existing patches, which are then placed in regions of an image in order to have the highest impact on final predictions. Similar to the Wasserstein GAN training~\cite{arjovsky2017wasserstein}, the PSGAN discriminator is updated several times in each epoch, whereas the generator is updated only once per epoch. Before each update, a minibatch of images and patches is sampled. The given minibatch of patches is fed to the generator to create the adversarial patches. Moreover, an attention model is included to determine a patch position that has the highest impact on the class prediction. %This is performed on each image of the minibatch, followed by placement of the adversarial patch \(\delta^i\) onto the image \(x^i\) via masking, thus creating the adversarial image. 

%The PSGAN generator is CNN with the U-Net architecture~\cite{ronneberger2015unet}, which outputs a segmentation of the image. The PSGAN discriminator is a CNN built for image classification.   %It consists of an encoder, that performs a feature contraction by increasing the number of channels in each layer, and a decoder part, which enlarges the feature map reducing the number of channels with each layer. 

%In the PSGAN training, the discriminator is updated using  the same objective as PhysGAN and the standard GAN algorithm, which is the \(L_{GAN}\), for a fixed generator. The generator, however, takes, besides \(L_{GAN}\) two other loss functions into consideration during its update.  Firstly, the \(L_{patch}\) loss constrains the perturbation added to the initial patch as it describes the difference between the pixel values of the adversarial patch and the initial patch. This difference is then minimized by the generator as its parameters are updated via gradient descent. Secondly, \(L_{adv}\)is the loss function of the model under attack. The generator learns to create adversarial attacks in order to minimize the loss of the targeted model for its generated adversarial images.

Closely related to PSGAN is the \textit{Inconspicuous Adversarial Patches (IAP)} framework~\cite{bai2021mobile}, which replicates the process of patch generation in PSGAN and repeats it for a series of generator-discriminator pairs. The goals is thus to reduce the conspicuousness of the patch by feeding it through the chain of GAN models. In the beginning, the background images are analyzed and an attention map indicating the best position for patch placement is calculated. Each GAN pair represents a step in the coarse-to-fine patch creation as it takes in the patch and background image at a different scale. The GAN training process remains the same as the generator aims to create realistic patches while the discriminator tries to distinguish them from the original images. IAP-generated patches aim to be indistinguishable from the background and thus resemble transparent masks.

\subsubsection{Using a Pretrained Generator}
In the second category, no full GAN training is performed. Instead, a pretrained GAN is used to improve patch appearance. The \emph{Naturalistic Physical Adversarial Patch Attack}, developed by Hu et al.~\cite{hu2021naturalistic}, aims to optimize for an adversarial patch in the GAN latent space directly. First, the patch is initialized as a noise vector. After the initialization, it performs a gradient update for each epoch and for each image, on which the patch is placed before the attack. For each iteration, the noise is fed to the generator to obtain the adversarial patch. The resulting patch is then added to the current image, which is then passed to the object detector. To perform an attack, a bounding box with the highest objectness probability and highest class probability is selected. The gradient descent is then used on the resulting loss, which also contains a total variation loss. Using the approach described above, Hu et al. performed several digital attacks, where they experimented with six different patch sizes, as well as physical attacks.

\emph{Universal NaTuralistic adversarial paTches (TnT) attack}~\cite{doan2021tnt} is another approach relying on a pretrained GAN. This approach aims at attacking image classifiers with realistic universal patches. It uses Wasserstein GAN~\cite{arjovsky2017wasserstein} with gradient penalty, which was pretrained on a dataset of flower images. For the background images, they used images from the ImageNet dataset to test the effectiveness of the attack in white-box and black-box setting. The TnT attack with high confidence scores on the pretrained image classifier had an up the three times higher attack success rate than the laVAN patch attack~\cite{karmon2018laVAN} while testing on the same image set. Finally, Doan et al. managed to create adversarial patches that resemble flowers, thus being less attention grabbing, but impacting the targeted classification model.

%The objective function is defined as the difference between the prediction score for the adversarial images and their corresponding targeted classes, \(L(I_{adv}, y_{target})\), and the prediction score for the adversarial images and their corresponding source or ground-truth classes, \(L(I_{adv}, y_{source})\). The second component describes the objective of an untargeted attack and, according to the authors, it speeds up the convergence of the attack. They apply gradient descent to update the latent noise vector \(z\) so as to minimize the total loss \(L_{total}\) of the model under attack when it analyzes the adversarial images \(I_{adv}\). 

%To generate a universal patch, which can be used to attack any image in the dataset, Doan et al. apply the same patch to a batch of images and feed the resulting adversarial images to the model under attack. They calculate the highest scoring class \(y_{max}\) for the current adversarial image \(img^i_{adv}\). If the highest scoring class \(y_{max}\) differs from the original label \(y_{source}\) or is actually the targeted class \(y_{target}\), then it is considered a successful attack. Doan et al. set a threshold \(fool_{thresh}\) for the number of successful attacks performed during training. Once the threshold is reached, the patch attack is considered as converged and \emph{universal}. 

\section{Approach}
We identify two major groups of GAN-based approaches to generate inconspicuous patches and describe the proposed pipelines, adapted for the object detection use case. Our pipeline assumes using a white-box gradient-based approach for adversarial patch generation.

\subsection{Combined Patch-GAN Training}
In the first approach we incorporate adversarial patch training directly into the GAN training pipeline. This method attempts to map the processes of PhysGAN~\cite{kong2020physgan} and PSGAN~\cite{liu2019psgan} models from steering model prediction and image classification respectively to the object detector attack. We thus simultaneously train a GAN model to create a latent space of realistic-looking patches and an adversarial patch to deter the object detector.

\begin{figure}[h]
	\centering
	\includegraphics[width=0.48\textwidth]{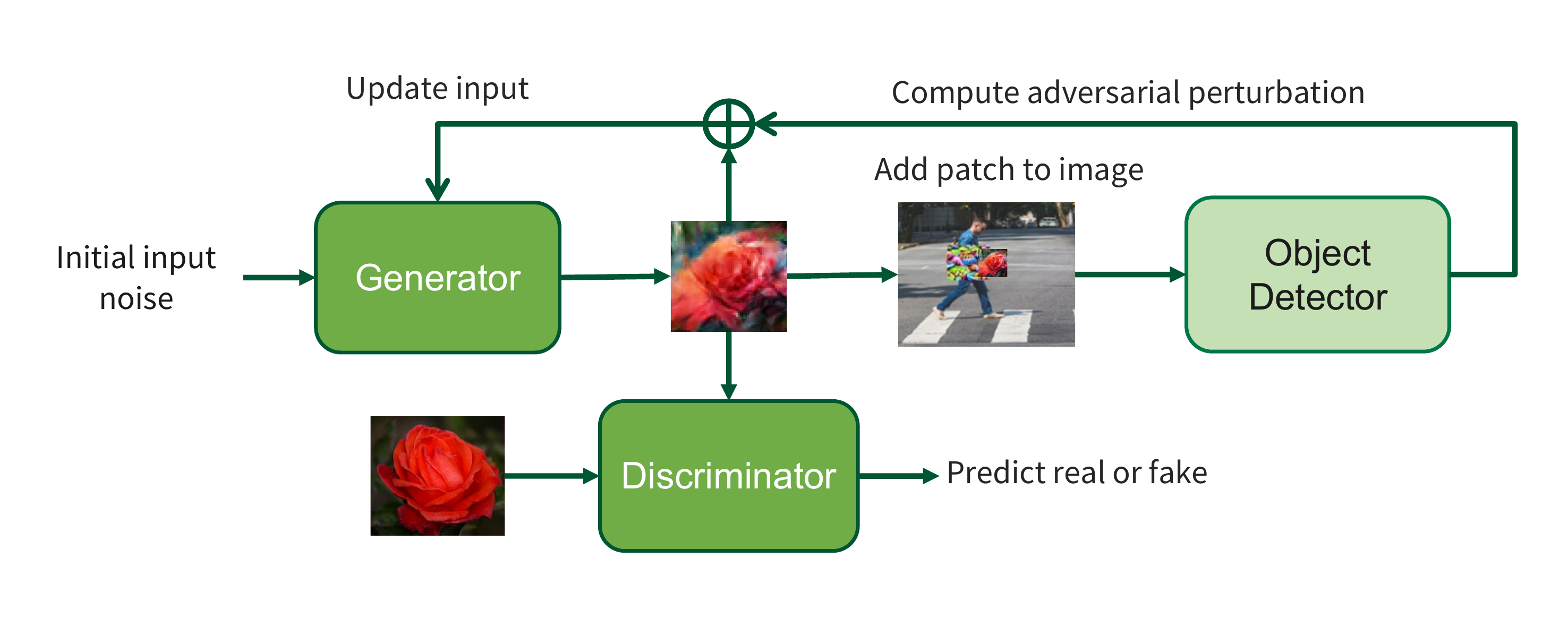}
	\caption{Overview of the combined Patch-GAN training}
	\label{fig:ganpatch}
\end{figure}

An overview of the training pipeline in the case of the \textit{combined Patch-GAN} attacks is presented in Figure~\ref{fig:ganpatch}. The patch is initialized randomly in the generator input format and undergoes two updates in each training epoch: one after the GAN training phase and one after the loss computation of the targeted object detector. Updating the patch after a GAN training step aims to restrict the patch to the latent space of realistic images developed by the GAN model. %Figure~\ref{fig:ganpatch} exemplifies this in the left side of the diagram, where the generator learns to create flower images from the patch noise while the discriminator tries to distinguish the fake flower images from real dataset samples. The generated patch is added to a background image and fed to the targeted model. The adversarial perturbation, which strengthens the attack towards the object detector, is then obtained with the patch update after the loss of the model for the adversarial image is computed. 

We further consider two extensions to the algorithm. First, we introduce a second generator update over the detector loss for the adversarial predictions. It takes into consideration the GAN loss for the generator, which gets the current patch as an input, and the loss of the object detector for the adversarial image. The current patch generation approach differs from PSGAN as the GAN loss is computed only over the patch and not over the entire adversarial image, similar to the PhysGAN.

% \begin{equation}
% 	\nabla_{\phi} V(G_\theta) = L_{GAN} + l_M(x^i_{adv}) 
%     \label{PGD-GAN2:loss_total}
% \end{equation}

% where

% \begin{equation}
% 	L_{GAN} = \frac{1}{m} \nabla_\phi \sum[log(1 - D_\phi(G(\delta^i)))]
%     \label{PGD-GAN2:loss_gan}
% \end{equation}

Second, we use two different random noise vectors during the patch training. One noise vector is reinitialized with each epoch and background image as it is used to train the two GAN components, while the other is the actual patch noise, initialized as before and optimized with each epoch and background image with the goal of reducing the loss of the object detector under attack.

\subsection{Patch Generation using a Pretrained GAN}
\begin{figure}[h]
	\centering
	\includegraphics[width=0.48\textwidth]{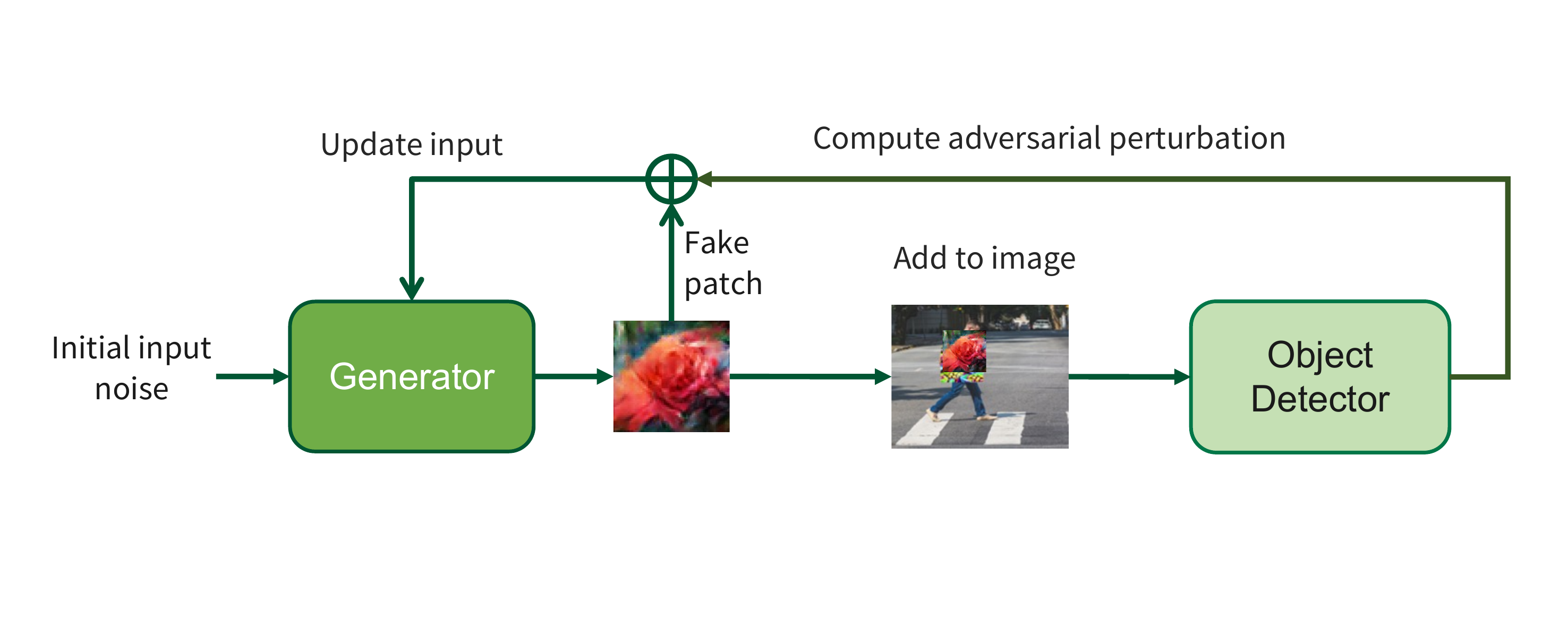}
	\caption{Overview of the patch training with a pretrained GAN generator}
	\label{fig:pretrained}
\end{figure}

The second approach focuses on restricting the trained patch to the images generated by a previously trained GAN model. Figure~\ref{fig:pretrained} shows the simplified pipeline for a \emph{Pretrained GAN Patch Attack}. In this approach, random noise is fed into the generator to obtain a realistic image. Similar to the combined Patch-GAN approach, the patch is applied to a background image and the resulting adversarial image is passed to the object detector under attack. The patch is then optimized to change the loss of the object detector. However, the parameters of the generator are no longer updated during patch training as in the previous approach. 

Our approach differs from the \emph{Naturalistic Physical Patch Attack}~\cite{hu2021naturalistic} in the attack procedure. In particular, we no longer target the single \textit{person} class and also focus on considering all objects in the image instead of a single object having the highest objectness and class probabilities. % We also  The current approach samples a noise vector from the input distribution for the selected generator. The noise is then used to create the patch by feeding it to the pretrained generator. After placing the adversarial patch on the current background image, the detector loss and the total variation component are calculated. 

\section{Experiments and Evaluation}
To evaluate the feasibility of the identified GAN-based approaches for inconspicuous patch generation for the object detection use case, we run experiments using YOLOv3~\cite{redmon2018yolov3} as a model under attack.

\subsection{Dataset and Models}
We have performed experiments with  YOLOv3 model~\cite{redmon2018yolov3}, using an open source Python implementation\footnote{\href{https://github.com/eriklindernoren/PyTorch-YOLOv3}{https://github.com/eriklindernoren/PyTorch-YOLOv3}}, detection was performed at the resolution 416x416 pixels.

\begin{figure}[t]
	\centering
    \begin{subfigure}{.485\linewidth}
    	\includegraphics[width=\textwidth]{fig/000000557923.jpg}
    	\caption{Input image}
    \end{subfigure}
     \begin{subfigure}{.485\linewidth}
    	\includegraphics[width=\textwidth]{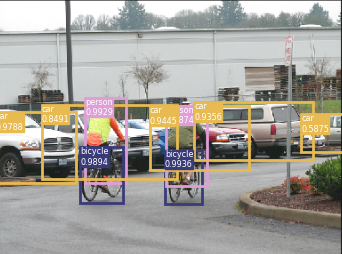}
    	\caption{YOLOv3 predictions}
    \end{subfigure}
    \caption{YOLOv3 predictions on an unattacked COCO image for the per-instance experiments}
    \label{fig:clean-img}
\end{figure}

The images to be attacked come from the COCO dataset~\cite{lin2015microsoft}. For the per-instance attacks, we use an exemplary COCO image (see Figure~\ref{fig:clean-img}).

We use two GAN architectures: DCGAN~\cite{radford2016unsupervised} and BigGAN~\cite{brock2018biggan}. To train DCGAN, we have used the Flower Recognition dataset~\cite{flowers-recognition}. The dataset contains 4,242 flower images of 320x240 pixels equally split into the classes \textit{daisy, dandelion, rose, sunflower, and tulip}. The dataset was built for image classification, not for unsupervised training for image generation as needed for the GAN models. Therefore, we performed dataset cleaning by manually removing the images containing scenarios such as a field of flowers or humans holding flowers, as these represent outliers from the intended GAN latent space, namely single flower generation. The clean Flowers Recognition dataset thus contains 1,385 images.

DCGAN was trained with the batch size of 64 with the Adam optimizer and learning rate 0.0002. The generated images have a size of 64x64 pixels and are further resized to reach the patch size. For BigGAN, we used the open source PyTorch re-implementation\footnote{\href{https://github.com/huggingface/pytorch-pretrained-BigGAN}{https://github.com/huggingface/pytorch-pretrained-BigGAN}}, pretrained on Imagenet.

We use the PGD algorithm~\cite{madry2017towards} for attacks. All trainings were performed on an NVIDIA RTX 1080 Ti GPU with 11GB VRAM.

\subsection{Conspicuous Baseline Patches}
To enable a fair comparison, we have first generated conventional adversarial patches using PGD. We have focused on the object vanishing attack, i.e. we have applied loss maximization using empty ground truth labels to enforce suppression of object detections. 

\begin{figure}[t]
	\centering
    \begin{subfigure}{0.485\linewidth}
    	\includegraphics[width=\textwidth]{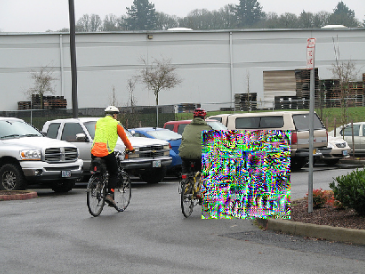}
    	\caption{Patch size 100x100 pixels, 7K epochs, lr=0.01}
    	\label{fig:pgd_no_augm_1}
    \end{subfigure}
    \begin{subfigure}{0.485\linewidth}
    	\includegraphics[width=\textwidth]{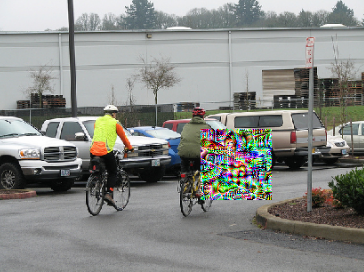}
    	\caption{Patch size 80x80 pixels, 5K~epochs, lr=0.5}
    	\label{fig:pgd_no_augm_2}
    \end{subfigure}
    
    \begin{subfigure}{0.485\linewidth}
    	\includegraphics[width=\textwidth]{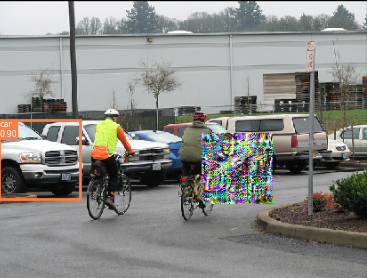}
    	\caption{Patch size 80x80 pixels, 15K epochs, lr=0.01}
    	\label{fig:pgd_no_augm_not_working1}
    \end{subfigure}
    \begin{subfigure}{0.485\linewidth}
    	\includegraphics[width=\textwidth]{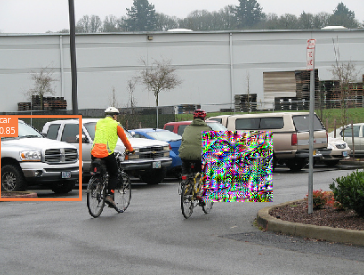}
    	\caption{Patch size 80x80 pixels, 15K~epochs, lr=0.02}
    	\label{fig:pgd_no_augm_not_working2}
    \end{subfigure}
    
    \begin{subfigure}{0.485\linewidth}
    	\includegraphics[width=\textwidth]{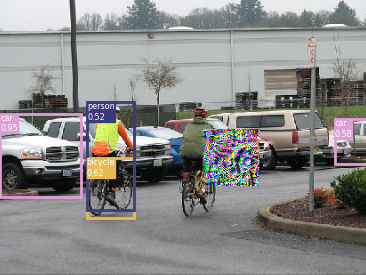}
    	\caption{Patch size 60x60 pixels, 10K epochs, lr=0.02}
    	\label{fig:pgd_no_augm_not_working3}
    \end{subfigure}
    \begin{subfigure}{0.485\linewidth}
    	\includegraphics[width=\textwidth]{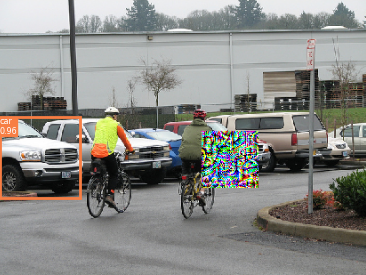}
    	\caption{Patch size 60x60 pixels, 10K epochs, lr=0.5}
    	\label{fig:pgd_no_augm_3}
    \end{subfigure}
    \caption{Attacks with conventional PGD patches}
    \label{fig:pgd_no_augm}
\end{figure}

Figure~\ref{fig:pgd_no_augm} demonstrates the PGD patches of different sizes, We have experimented with various training times and learning rates. The 100x100 pixels PGD patch requires 7K epochs at learning rate 0.01 to suppress all bounding boxes (see Figure ~\ref{fig:pgd_no_augm_1}). The 80x80 pixels patch only achieves the same result in 5K epochs when using a learning rate of 0.5 as seen in Figure~\ref{fig:pgd_no_augm_2}. Training of the 80x80 pixels patch with learning rates of 0.01 and 0.02 did not manage to suppress all bounding boxes even after 15K epochs (see Figures~\ref{fig:pgd_no_augm_not_working1} and~\ref{fig:pgd_no_augm_not_working2}). Because of its smaller attack surface, we train the 60x60 pixels patch directly with a learning rate of 0.02. Figure~\ref{fig:pgd_no_augm_not_working3} shows, however, that this patch does not manage to suppress four bounding boxes, which are placed towards the image margins. Using the learning rate of 0.5 for the 60x60 pixels patch gives better results. Only one bounding box remains in Figure~\ref{fig:pgd_no_augm_3}. The performance of the 60x60 pixels patch stagnates and the confidence score of the remaining bounding box does not decrease after 100K epochs, at which point the training is stopped.

Overall, the conventional PGD patches are able to completely supress all detections in an input images using a sufficiently large patch (at least 80x80x pixels, i.e. 3\% of an input). The smaller the patch, the more it profits from a higher learning rate and longer training time.

%For the further experiments, we focus on 100x100 pixels patches.

\subsection{Combined PGD-GAN Training}
For the combined PGD-GAN approach, we used the DCGAN architecture, while the PGD attack was implemented as done for the conspicuous baseline. 

Following the baseline, a model with one discriminator and generator update per training step was first evaluated. After generator and discriminator parameters are updated at step $t$, the generator gets an updated patch at step $t+1$ and outputs a new patch. We then insert the new patch in the COCO image and produce predictions with the YOLOv3 object detector. After computing the YOLOv3 loss, the patch optimizer was run in order to update the current patch state. This approach led to highly distorted patches not resembling the dataset, whereas the patch itself had no impact on the surrounding bounding boxes.

\begin{figure}[t]
	\centering
    \begin{subfigure}{0.485\linewidth}
    	\includegraphics[width=\textwidth]{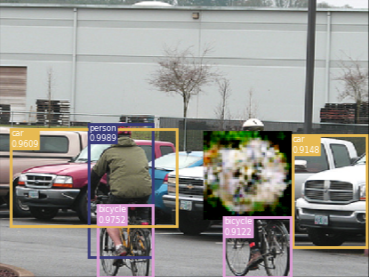}
    	\caption{Cropping and horizontal flipping, 2K epochs}
    	\label{fig:PGD-GANv2_2000}
    \end{subfigure}
    \begin{subfigure}{0.485\linewidth}
    	\includegraphics[width=\textwidth]{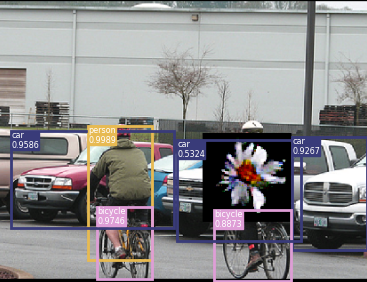}
    	\caption{Cropping and horizontal flipping, 4K epochs}
    	\label{fig:PGD-GANv2_4000}
    \end{subfigure}
    \begin{subfigure}{.485\linewidth}
    	\includegraphics[width=\textwidth]{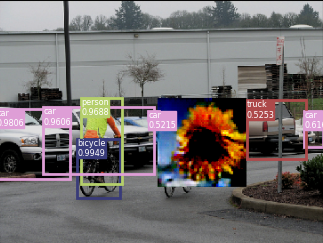}
    	\caption{2,5K epochs}
    	\label{fig:PGD-GAN_v2_2}
    \end{subfigure}
    \begin{subfigure}{.485\linewidth}
    	\includegraphics[width=\textwidth]{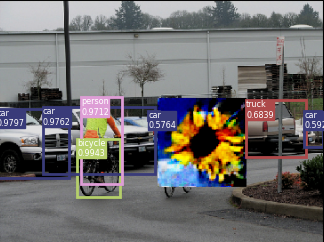}
    	\caption{5K epochs}
    	\label{fig:PGD-GAN_v2_3}
    \end{subfigure}
    \caption{Attacks with the combined PGD-GAN training using two generator updates per epoch}
    \label{fig:PGD-GAN_v2}
\end{figure}

We have achieved better results via introducing a second generator update. We thus updated generator twice per epoch: first during the GAN training step and then after the patched image is evaluated and the loss of YOLOv3 is calculated. Figure~\ref{fig:PGD-GANv2_4000} shows, that, the patch images remain in the dataset distribution after 4K epochs. However, with each newly generated image, a different flower type is created (a dandelion at 2K epochs and a daisy at 4K epochs). At both stages the covered person is not detected and the confidence score for the car in the back decreases.

\begin{figure}[t]
	\centering
    % \begin{subfigure}{.485\linewidth}
    % 	\includegraphics[width=\textwidth]{fig/combined/v3_122734_pgd_dc_gan_patch_clean_cropped.png}
    % 	\caption{Clean image evaluation}
    % 	\label{fig:PGD-GANv3_clean}
    % \end{subfigure}
    % \begin{subfigure}{.485\linewidth}
    	\includegraphics[width=.24\textwidth]{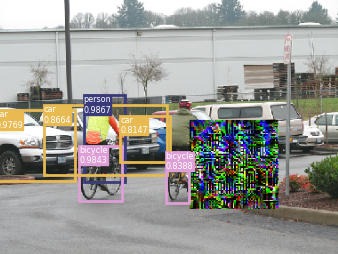}
    % 	\caption{After 1K epochs}
    % 	\label{fig:PGD-GANv3_1000}
    % \end{subfigure}
    \caption{Attack with the combined PGD-GAN training with generator update after the discriminator and patch updates. Results are shown after 1K epochs}
    \label{fig:PGD-GANv3}
\end{figure}

Figures~\ref{fig:PGD-GAN_v2_2} and \ref{fig:PGD-GAN_v2_3} demonstrate how this version of the algorithm performs without horizontal flipping. The patch covers an entire cyclist, which prevents it from being identified. Moreover the adjacent cars are identified as such only with a 0.57 and 0.68 confidence score respectively, which are lower than in the corresponding clean image. However, the confidence score does not decline linearly over the epochs. For instance, the red bounding box in Figure~\ref{fig:PGD-GAN_v2_3} displays a higher confidence score of 0.68 at epoch 5K compared to only 0.52 in epoch 2500 as shown in Figure~\ref{fig:PGD-GAN_v2_2}.

The last model that we have evaluated included updating the generator once per epoch, after both the discriminator and the patch were updated. Figure~\ref{fig:PGD-GANv3} shows that the patch developed with this method manages to suppress more bounding boxes in the neighbouring region. However, the generator obviously does not learn the distribution of the GAN training dataset. 

% We also provide patches generated with both versions of the algorithm in Figure~\ref{fig:gan_comp_1000}. The generator trained with the version 2 algorithm is developing images similar to the training dataset distribution after 1K epochs already, while the generator trained during the version 3 algorithm only creates noise pattern similar to the conventional PGD patches. 

% \begin{figure}[t]
% 	\centering
%     \begin{subfigure}{\linewidth}
%     	\centering
%     	\includegraphics[width=\textwidth]{fig/combined/20220319-174022_pgd_dc_gan_patch_clean_resize_augmentation_empty_1000ep_gan.png}
%     	\caption{Generator output after 1K epochs of version 2 algorithm}
%     	\label{fig:gan_v2_1000}
%     \end{subfigure}
%     \begin{subfigure}{\linewidth}
%     	\centering
%     	\includegraphics[width=\textwidth]{fig/combined/v3_122734_pgd_dc_gan_patch_1000ep_gan.png}
%     	\caption{Generator output after 1K epochs of version 3 algorithm}
%     	\label{fig:gan_v3_1000}
%     \end{subfigure}
%     \caption{Patches generated during combined GAN-PGD training}
%     \label{fig:gan_comp_1000}
% \end{figure}

In summary, we could generate realistic looking adversarial patches with the combined approach. The best performing version of the algorithm included two generator updates per epoch. The attack success, however, is worse than when conspicuous adversarial patches are used.

\subsection{Using a Pretrained Generator}
Next, we evaluate the usage of a pretrained GAN generator. We have experimented with two GAN models: DCGAN and BigGAN. DCGAN was trained for 2K epochs on the Flowers Recognition dataset, which was preprocessed as described above.  For the BigGAN, we have used the pretrained model and set the chosen class to daisy (985). The patch optimization is the same as for the conspicuous baseline, the weight for the total variation  is set to 0.01. 

\begin{figure}[t]
	\centering
    % \begin{subfigure}{0.485\linewidth}
    % 	\centering
    % 	\includegraphics[width=\textwidth]{fig/pretrained/20220323-014252_pretrained_dc_clean - Copy.png}
    % 	\caption{Clean image evaluation}
    % 	\label{fig:pretr_interp1}
    % \end{subfigure}
    \begin{subfigure}[t]{0.485\linewidth}
    	\centering
    	\includegraphics[width=\textwidth]{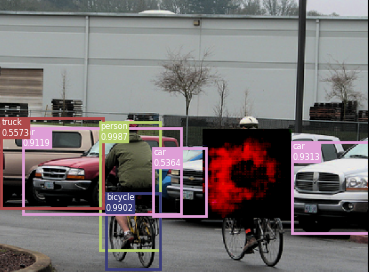}
    	\caption{Interpolation, augmentations, 1K epochs}
    	\label{fig:pretr_interp2}
    \end{subfigure}
    \begin{subfigure}[t]{0.485\linewidth}
    	\centering
    	\includegraphics[width=\textwidth]{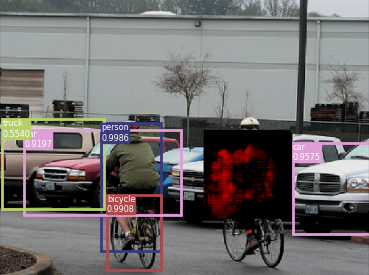}
    	\caption{Interpolation, augmentations, 3K epochs}
    	\label{fig:pretr_interp3}
    \end{subfigure}
        \begin{subfigure}[t]{0.485\linewidth}
    	\centering
    	\includegraphics[width=\textwidth]{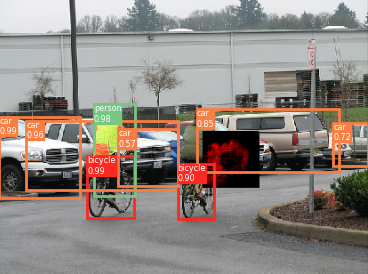}
    	\caption{No interpolation, 3K epochs}
    	\label{fig:pretr_nointerp2}
    \end{subfigure}
    \begin{subfigure}[t]{0.485\linewidth}
    	\centering
    	\includegraphics[width=\textwidth]{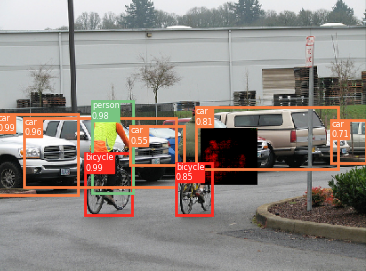}
    	\caption{No interpolation, 5K epochs}
    	\label{fig:pretr_nointerp3}
    \end{subfigure}
        \begin{subfigure}[t]{0.485\linewidth}
    	\centering
    	\includegraphics[width=\textwidth]{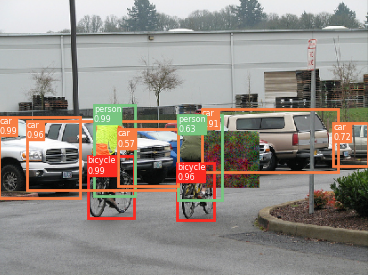}
    	\caption{With latent shift applied, 1K epochs}
    	\label{fig:latent_shift}
    \end{subfigure}
    \begin{subfigure}[t]{0.485\linewidth}
    	\centering
    	\includegraphics[width=\textwidth]{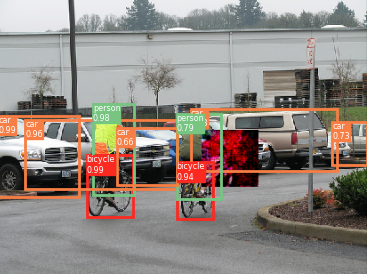}
    	\caption{With patch transformations applied, 5K epochs}
    	\label{fig:transform}
    \end{subfigure}
    \caption{Attacks with a pretrained DCGAN}
    \label{fig:pretr_dcgan}
\end{figure}

Figure~\ref{fig:pretr_dcgan} demonstrates the results for the experiments with the pretrained DCGAN. We have first experimented with patches resized from 64x64 as generated by DCGAN to 100x100 pixels using interpolation. As Figure~\ref{fig:pretr_interp2} shows, the patch is placed on a cyclist, which deters the object detector from recognizing the person, the bicycle and the car behind them after 1K epochs using a learning rate of 0.01. Moreover, the car to the left of the patch has the reduced confidence score of 0.53 compared to the clean image score of 0.76. However, a major problem here is that the patch is getting darker with each training epoch (see Figure~\ref{fig:pretr_interp3}). %The following experiments renounce the use of interpolation for resizing the patch as well as the augmentation of the COCO image. All of the following patches are trained in the original size for both GAN models.

Experiments without patch interpolation (i.e. using patches of size 64x64 pixels as generated by DCGAN) also show the same darkening effect (see Figures~\ref{fig:pretr_nointerp2} and \ref{fig:pretr_nointerp3}). As expected, these patches also do not suppress the surrounding boxes as well as the previous experiment due to their smaller attack surface, but the confidence scores are decreased. This is also consistent with our conspicuous baseline experiments. The adversarial patch, generated with the pretrained generator, only covers part of the cyclist but the object detector cannot detect a person. In addition, the bounding boxes surrounding the patch have a lower confidence score. Affected are the detections of the cars to the right of the patch as well as the bicycle below it.

To mitigate the darkening patch effect, we further evaluate two countermeasures. First, we apply latent shift interpolation. For that, we initialize a patch mask of randomly distributed values and then apply it to the patch via interpolation. This procedure is repeated during each training epoch before applying the patch to the COCO image. Figure~\ref{fig:latent_shift} shows results for this approach after 3K training epochs. In this case, the patch value does not remain in the DCGAN image distribution, but resembles noise, which diverges from the flower images, and does not improve with longer training time. Moreover, the patch performs worse than the previous experiments during the evaluation. The person to the left is recognized by the object detector albeit with a lower score than in the clean image. The other surrounding bounding boxes do not have a considerably reduced confidence score. 

% \begin{equation}
%     patch = (1 - patch)\ *\ mask + patch\ *\ mask
%     \label{eq:latent}
% \end{equation}

\begin{figure}[t]
	\centering
    \begin{subfigure}[t]{0.485\linewidth}
    	\centering
    	\includegraphics[width=\textwidth]{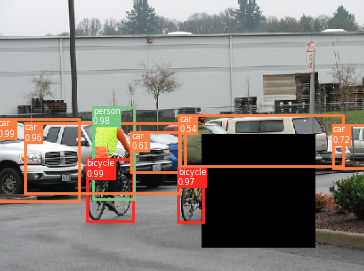}
    	\caption{Standard training, 7K epochs}
    	\label{fig:biggan_exp1}
    \end{subfigure}
    \begin{subfigure}[t]{0.485\linewidth}
    	\centering
    	\includegraphics[width=\textwidth]{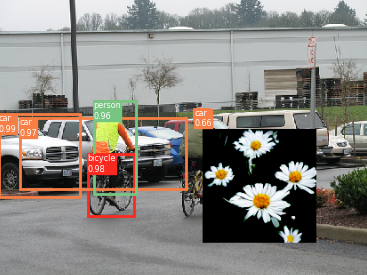}
    	\caption{With latent shift applied, 7K epochs}
    	\label{fig:biggan_exp2}
    \end{subfigure}
    \begin{subfigure}[t]{0.485\linewidth}
    	\centering
    	\includegraphics[width=\textwidth]{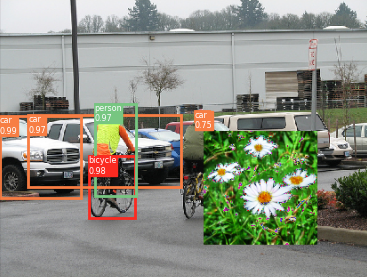}
    	\caption{With patch transformations applied, 7K epochs}
    	\label{fig:biggan_exp3}
    \end{subfigure}
    \begin{subfigure}[t]{0.485\linewidth}
    	\centering
    	\includegraphics[width=\textwidth]{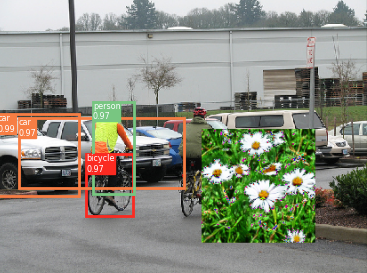}
    	\caption{With patch transformations applied, 10K epochs}
    	\label{fig:biggan_exp3_10K}
    \end{subfigure}
    \caption{Attacks with a pretrained BigGAN}
    \label{fig:biggan_exp}
\end{figure}

A further attempt, aiming to improve the appearance of patches, is the usage of patch transformation, as suggested in ~\cite{thys2019fooling}. This approach aims at making patches more robust and includes a number of transformations applied to a patch before it is added to an input image. In includes adding random noise to the patch as well as random changes in patch brightness and contrast. In particular, we first multiply the patch with a contrast mask and then add brightness and noise masks. In all cases, masks include randomly sample values, the contrast interval is restricted to [0.8, 1.2], the brightness interval is restricted to [-0.1, 0.1], the noise mask contains values in the interval [-0.1, 0.1]. As can be seen in Figure~\ref{fig:transform}, the patch stays in the latent space of the DCGAN model after 5K epochs. This, however, comes at a cost of small rise in the confidence of object detections near the patch. %However, the person to the left is identified with a higher score of 0.79. The car to the left is also identified with a higher confidence of 0.66, compared to the previous patch attacks. The other bounding boxes show no considerable changes.

As the figures demonstrate, in our DCGAN experiments we have no control over the generated flower class, so that patches may contain various flowers during the training. 

\begin{figure}[t!]
	\centering
	\begin{subfigure}[t]{0.48\linewidth}
    	\includegraphics[width=\textwidth]{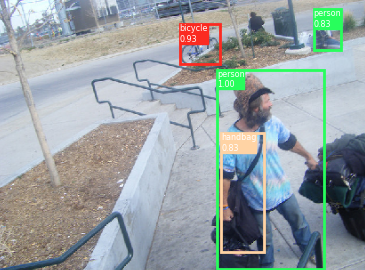}
    	\caption{No attack}
    	\label{fig:univ_dc_clean}
    \end{subfigure}
    \begin{subfigure}[t]{0.48\linewidth}
    	\includegraphics[width=\textwidth]{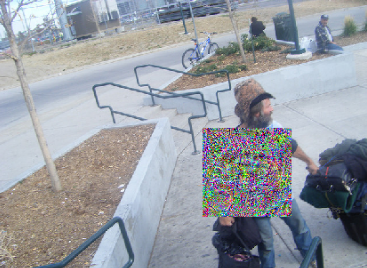}
    	\caption{Attack with a conspicuous PGD patch}
    	\label{fig:univ_dc_1000}
    \end{subfigure}
    
    \begin{subfigure}[t]{0.48\linewidth}
    	\includegraphics[width=\textwidth]{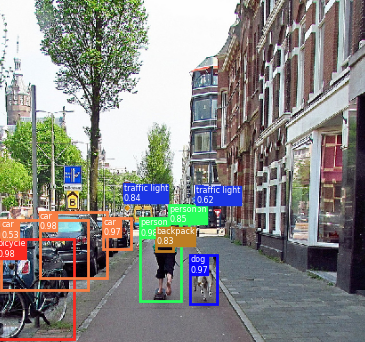}
    	\caption{No attack}
    	\label{fig:univ_dc_clean}
    \end{subfigure}
    \begin{subfigure}[t]{0.48\linewidth}
    	\includegraphics[width=\textwidth]{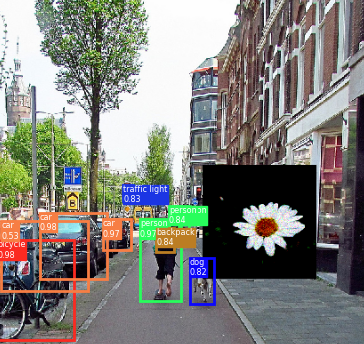}
    	\caption{Attack with a pretrained BigGAN with latent shift}
    	\label{fig:univ_dc_1000}
    \end{subfigure}
    
    \begin{subfigure}[t]{0.48\linewidth}
    	\includegraphics[width=\textwidth]{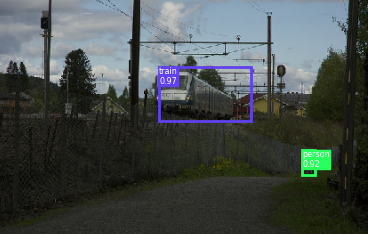}
    	\caption{No attack}
    	\label{fig:univ_big_clean}
    \end{subfigure}
    \begin{subfigure}[t]{0.48\linewidth}
    	\includegraphics[width=\textwidth]{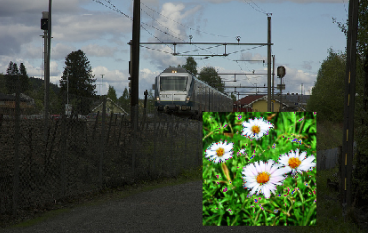}
    	\caption{Attack with a pretrained BigGAN with patch transformations}
    	\label{fig:univ_big_1000}
    \end{subfigure}
    
    \begin{subfigure}[t]{0.48\linewidth}
    	\includegraphics[width=\textwidth]{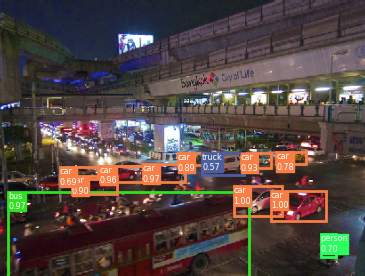}
    	\caption{No attack}
    	\label{fig:univ_big_clean}
    \end{subfigure}
    \begin{subfigure}[t]{0.48\linewidth}
    	\includegraphics[width=\textwidth]{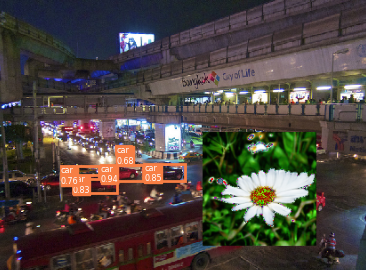}
    	\caption{Attack with a pretrained BigGAN with patch transformations}
    	\label{fig:univ_big_1000}
    \end{subfigure}
    
    \begin{subfigure}[t]{0.48\linewidth}
    	\includegraphics[width=\textwidth]{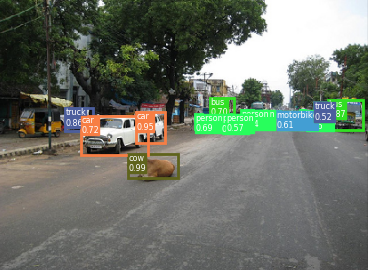}
    	\caption{No attack}
    	\label{fig:univ_dc_clean}
    \end{subfigure}
    \begin{subfigure}[t]{0.48\linewidth}
    	\includegraphics[width=\textwidth]{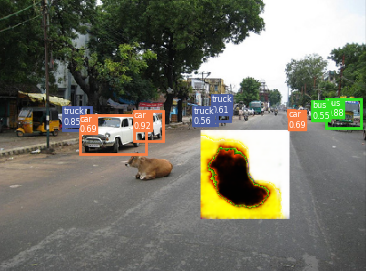}
    	\caption{Attack with a pretrained DCGAN with patch transformations}
    	\label{fig:univ_dc_1000}
    \end{subfigure}
    
    \caption{Examples of universal patch attacks generated for a subset of COCO, all trained for 1K epochs}
    \label{fig:univ_big_train}
\end{figure}

We further repeat the experiments with the BigGAN model. The chosen class is 985 representing daisies. The experiments are performed with the patches of 128x128 pixels, which is the size of the original BigGAN generator output. Figure~\ref{fig:biggan_exp1} displays the patch attack result after 7K epochs. It turns completely black, however it still manages to suppress the identification of the person to the left of the patch. 

Next, we assess the effect of adding the interpolation with the latent value. Figure~\ref{fig:biggan_exp2} shows the patch resulting from 7K epochs. In this case, only the background of the flower images turns black while the flowers remain clearly visible. Moreover, the patch manages to suppress the bounding boxes of the cars above and to the right of its position as well as the identification of the first cyclist and the first bicycle on the left.

Finally, we apply patch transformations. This helps to fully overcome the problem of the dark patch background, as the patch background is not longer black, but resembles a field. As Figure~\ref{fig:biggan_exp3} shows, the patch achieves similar results to the previous BigGAN experiment from Figure~\ref{fig:biggan_exp2}. It suppresses the same bounding boxes and shows a confidence score of 0.75 for the car bounding box in the upper left corner of the patch. This score is higher than in the previous BigGAN experiment but lower than in the clean image. Moreover, by training the pretrained BigGAN patch with transformation for 10K epochs on one COCO image, the bounding box in the upper left corner is suppressed as well (see Figure \ref{fig:biggan_exp3_10K}).

In summary, the approach involving a pretrained generator leads to a significantly higher image fidelity. In a standard setting, the patch tends to get completely black, but the proposed latent shift and patch transformations help to overcome the problem. As expected, BigGAN led to significantly better patches due to larger capacity.

\begin{table*}[t]
    \centering
    \begin{tabular}{l|c|cccccccc}
          & \rotatebox{90}{No attack} & \mybox{Black} & \mybox{Conventional PGD} & \mybox{\cite{hu2021naturalistic} with class=daisy} & \mybox{Pretrained BigGAN + latent shift} & \mybox{Pretrained BigGAN + transformations} & \mybox{Pretrained BigGAN + transformations} & \mybox{Pretrained DCGAN + transformations} & \mybox{Combined PGD-GAN training}  \\
          
          &  & \includegraphics[width=0.081\textwidth]{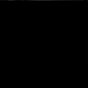} 
          & \includegraphics[width=0.081\textwidth]{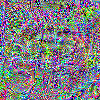} 
          & \includegraphics[width=0.081\textwidth]{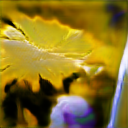} 
          %& \includegraphics[width=0.065\textwidth]{fig/universal/naturalistic-pug-images-1000.png} 
          & \includegraphics[width=0.081\textwidth]{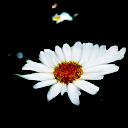} 
          & \includegraphics[width=0.081\textwidth]{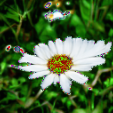} 
          & \includegraphics[width=0.081\textwidth]{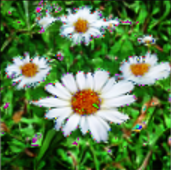} 
          & \includegraphics[width=0.081\textwidth]{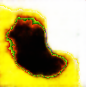} 
          & \includegraphics[width=0.081\textwidth]{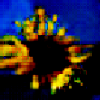} \\
         \hline
         mAP & 43.8 & 4.2 & \textbf{3.4} & 3.8 &  3.8 & 4.4 & 4.1 & 3.8 & 3.9 \\
         AP$_{person}$ & 80.3 & 55.2 & \textbf{28.4} & 40.4 & 41.8 & 39.3 & 32.0 & 41.2 & 46.0\\
         AP$_{bicycle}$ & 2.2 & \textbf{0.0} & \textbf{0.0} & \textbf{0.0} & \textbf{0.0} & 0.2 & 0.2 & \textbf{0.0} & \textbf{0.0}\\
         AP$_{car}$ & 7.9 & 3.1 & \textbf{0.0} & 1.2 & 0.3 & 0.4 & 0.3 & 0.3 & 0.3\\
        %  AP$_{bicycle}$ & 2.2 & 7.9 \\
        %  AP$_{car}$
        %  Black &  &  & 55.2 & 0 & 3.1\\
        %  PGD &  &  &  & 0 & 0 \\
        %  Naturalistic &  &  & 40.4 & 0 & 1.2\\
        %  Naturalistic &  &  & 40 & 0 & 1.3\\
        %  \hline
        %  Pretrained &  &  & 41.8 & 0 & 0.3\\
        %  BigGAN & & & & & \\
        %  Pretrained &  &  & 39.3 & 0.2 & 0.4  \\
        %  BigGAN & & & & & \\
        %  (transf.) & & & & & \\
        %  Pretrained &  &  &  & 0.2 & 0.3  \\
        %  BigGAN & & & & & \\
        %  (transf.) & & & & & \\
        %  Pretrained &  &  &  & 0 & 0.3 & \\
        %  DCGAN & & & & & \\
        %  (transf.) & & & & & \\
        %  Combined &  &  &  & 0 & 0.3 \\
        %  PGD-GAN & & & & & \\
    \end{tabular}  
    \caption{Mean average precision (mAP) and average precision (AP) for certain classes in \% for universal patches, generated for a subset of COCO}
    \label{tab:map}
\end{table*}

\subsection{Universal Inconspicuous Patches}
Finally, we evaluate whether the studied approaches to generate inconspicuous patches can also be applied in a universal manner. The goal of a universal attack is to fool all images with a single perturbation \cite{moosavi2017universal}. 

For the experiments, we create a subset of the COCO dataset, containing objects of classes \textit{person, car, bicycle}. The resulting subset contains 1,146 images, which are further split according to the COCO protocol to 1,101 train and 45 test images. All universal patch training experiments are run for 1K epochs over the entire training dataset. The patch learning rate is set to 0.01 and the GAN learning rate for the combined PGD-GAN patch attack is set to 0.0002. The patch size during training is set to the original size of the GAN architecture output (i.e., 64x64 for DCGAN and 128x128 for BigGAN) to avoid information loss through resizing. The patch placement is fixed similar in the per-instance experiments.

% \begin{figure}[htb]
% 	\centering
%     \begin{subfigure}{0.485\linewidth}
%     	\includegraphics[width=\textwidth]{fig/universal/pgd_patch_clean_cropped.png}
%     	\caption{Clean image}
%     	\label{fig:univ_pgd_clean}
%     \end{subfigure}
%     \begin{subfigure}{0.485\linewidth}
%     	\includegraphics[width=\textwidth]{fig/universal/pgd_patch_cropped.png}
%     	\caption{After 1K epochs}
%     	\label{fig:univ_pgd_1000}
%     \end{subfigure}
%     \caption{Examples of universal PGD patch attacks }
%     \label{fig:univ_pgd_train}
% \end{figure}

Using the conventional PGD patches, we could suppress all bounding boxes in the test images. The universal patch generated using the pretrained BigGAN with patch transformations for brightness and contrast was also successful (see Figure~\ref{fig:univ_big_train}). In comparison, the pretrained DCGAN patch attack has a reduced effect on the object detection (see Figure \ref{fig:univ_dc_1000}). However, it reduces the confidence scores of the surrounding bounding boxes significantly. One major difference to the previous example is the quality of the image and of the generated object respectively. The daisy image in this case is distorted and no longer recognizable as a flower.

\begin{figure}[h]
	\centering
    \includegraphics[width=0.25\textwidth]{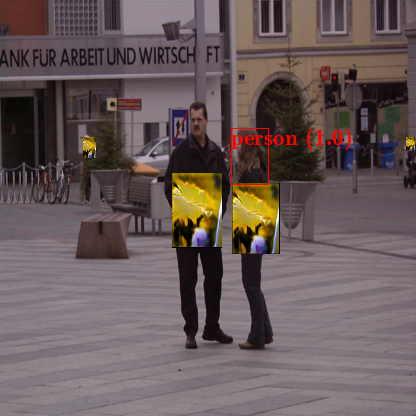}
    \caption{A naturalistic patch, created using the framework by Hu et al., attacks objects of the  class \textit{person} via overlapping with the clothing area}
    \label{fig:naturalistic-placement}
\end{figure}

We have trained and evaluated several patches using the same settings (see Table \ref{tab:map}). We have also evaluated a patch, generated using the approach by Hu et al.~\cite{hu2021naturalistic} using the open-source code\footnote{\href{https://github.com/aiiu-lab/Naturalistic-Adversarial-Patch}{https://github.com/aiiu-lab/Naturalistic-Adversarial-Patch}}. Following the procedure in the paper, the training was performed on the INRIA dataset~\cite{inria} for 1K epochs. We also set the class to daisy. Note, that direct comparison with the method by Hu et al.~\cite{hu2021naturalistic} is not possible due a different method to add patch to an image (see Figure \ref{fig:naturalistic-placement}). Instead of attacking object of a certain class by direct overlapping with a patch, we focus on a single patch at a fixed position in an image, which can attacks all objects.

Every approach managed to reduce the average mAP drastically, whereas the best result was obtained with the conventional PGD attack, as expected. The patch generated according to \cite{hu2021naturalistic} achieves the same mAP, as the pretrained BigGAN without transformations and the pretrained DCGAN with transformations. This patch also has the best results for the class \textit{person}, but worst for the class \textit{car}. Finally, the pretrained BigGAN patch with transformation scores the highest mAP for both images, being least effective overall. In the case of one of the pretrained BigGAN patches with patch transformations, the mAP score of 4.4\% is even higher than the black square mAP value. The patches generated with the pretrained BigGAN demonstrate, however, the most naturalistic appearance out of all universal experiments, also compared to the results obtained with the framework by Hu et al.~\cite{hu2021naturalistic}.

\section{Conclusion}
In this work, we have evaluated the existing GAN-based methods for inconspicuous patch generation on the object detection use case. Following the analysis of the state of the art, we have identified two groups of promising approaches: the first method focuses on combining the GAN training process with the training of the adversarial patch, while the second one relies on a pretrained GAN model during the patch training process. For each group, we have adapted the procedure to attack the object detector and ran the experiments on YOLOv3 as a model under attack both in per-instance and universal settings using the COCO dataset. All attacks were performed using the PGD algorithm. Differently from the state of the art, we focused on suppressing objects in the direct proximity of a patch, which is also a realistic attacks scenario.

Our experiments have demonstrated, that using the pretrained GAN generator leads to adversarial patches of higher visual fidelity. Better performing BigGAN led to more realistic looking patches compared to DCGAN. However, since BigGAN training on ImageNet is resource consuming, we have performed the experiments on combined PGD-GAN training only with a DCGAN model. Evaluating the combined training approach with a GAN of larger capacity might lead to even better results. 

During evaluation of the universal attacks, we could observe an evident trade-off between the patch appearance and the attack performance. Our pretrained DCGAN and combined PGD-GAN have demonstrated attack performance comparable to the state-of-the-art approach by Hu et al~\cite{hu2021naturalistic}, although no direct comparison is possible because of different patch placement approaches. The pretrained DCGAN approach as well as the PGD GAN approach led to a better attack success than the pretrained BigGAN method during evaluation. Overall, the performance on the test set under attack was significantly lower than on the clean images. Although the attack strength of the conspicuous patches could not be reached, the studied approaches present a promising trade-off between the attack success and naturalistic appearance.  

%Further research can involve training the GAN-based patches with a different dataset. Another step forward in researching inconspicuous patch generation would be to experiment with trained patches in a physical setting by printing out the patches on clothing items or as advertisement posters. In addition, a second discriminator can be used in the PGD GAN approach to distinguish between the attacked and the clean images during training. The patch can then be updated once more with the goal of maximizing the loss of this second discriminator. Finally, in the pretrained GAN method, the discriminator from the chosen GAN model can be used to evaluate the patch appearance and incorporate the corresponding loss in the objective function of the training procedure.

\section*{Acknowledgement}
  The research leading to these results is funded by the German Federal Ministry for Economic Affairs and Climate Action within the project “KI Absicherung“ (grant 19A19005W) and by KASTEL Security Research Labs. The authors would like to thank the consortium for the successful cooperation.

\bibliographystyle{named}
\bibliography{literature}

\end{document}